\documentclass{article}
\usepackage{graphicx} 
\usepackage[a4paper, total={6in, 8in}]{geometry}
\usepackage{hyperref}
\title{Transforming organic chemistry research paradigms: moving from manual efforts to the intersection of automation and artificial intelligence}
\author{
  Chengchun Liu \\
  School of Materials Science and Engineering, Peking University, Beijing, China\\
  \\
  Yuntian Chen \\
  Ningbo Institute of Digital Twin, Eastern Institute of Technology, Ningbo, China\\
  \texttt{ychen@eias.ac.cn} \\
  \\
  Fanyang Mo \\
  School of Materials Science and Engineering, Peking University, Beijing, China\\
  AI for Science (AI4S)-Preferred Program, Shenzhen Graduate School, Peking University, Shenzhen, China \\
  \texttt{fmo@pku.edu.cn}
}

\date{November 26, 2023}

\begin{document}

\maketitle

\begin{abstract}
Organic chemistry is undergoing a major paradigm shift, moving from a labor-intensive approach to a new era dominated by automation and artificial intelligence (AI). This transformative shift is being driven by technological advances, the ever-increasing demand for greater research efficiency and accuracy, and the burgeoning growth of interdisciplinary research. AI models, supported by computational power and algorithms, are drastically reshaping synthetic planning and introducing groundbreaking ways to tackle complex molecular synthesis. In addition, autonomous robotic systems are rapidly accelerating the pace of discovery by performing tedious tasks with unprecedented speed and precision. This article examines the multiple opportunities and challenges presented by this paradigm shift and explores its far-reaching implications. It provides valuable insights into the future trajectory of organic chemistry research, which is increasingly defined by the synergistic interaction of automation and AI.
\end{abstract}
	

\section{Introduction}
\noindent Traditional chemical research is characterized by a paradigm in which chemists explore and refine reaction conditions through a process of trial and error, supported by theoretical calculations and numerical simulations for scientific validation and elucidation. Critical advances have profoundly influenced the course of modern organic chemistry. Among these pivotal breakthroughs is the successful synthesis of the complex quinine alkaloids to combat malaria, achieved by R. B. Woodward and W. E. Doering from Harvard University in 1944 \cite{1}. Then, in 1965, W. Kohn at the University of California, Santa Barbara and L. Sham at the University of California, San Diego made a major contribution to theoretical chemistry by formulating density functional theory \cite{2}, an essential tool for calculating the electronic structure of molecules. In 1970, J. Pople at Northwestern University revolutionized computational chemistry with the introduction of the GAUSSIAN program \cite{3}, a versatile tool for quantum chemical calculations that provides accurate predictions and insight into molecular properties and behavior. 
	
However, despite these advances in computational chemistry, there remains a significant trade-off between the accuracy and speed of calculations. This trade-off can be particularly prominent when attempting to predict the outcomes of complex organic reactions. For many complex systems, ensuring high accuracy can require computational methods that are computationally intensive and time-consuming. Conversely, methods that are faster may employ approximations that compromise the accuracy of the results. This challenge remains one of the primary reasons why organic chemistry often still necessitates a manual, trial-and-error approach, as direct and reliable prediction remains elusive for many reactions. The manual process is not without its pitfalls. It is fraught with the potential for human error, often leading to lengthy research timelines and the waste of valuable resources \cite{4}. In addition, theoretical calculations can sometimes lose critical information when dealing with complex systems due to mathematical approximations \cite{5}. As a result, these labor-intensive, monotonous, and repetitive tasks often form the research bottleneck and pose significant challenges to today's chemists. 
	
\begin{figure}[!t]
		\centering
		\includegraphics[width=170mm, keepaspectratio]{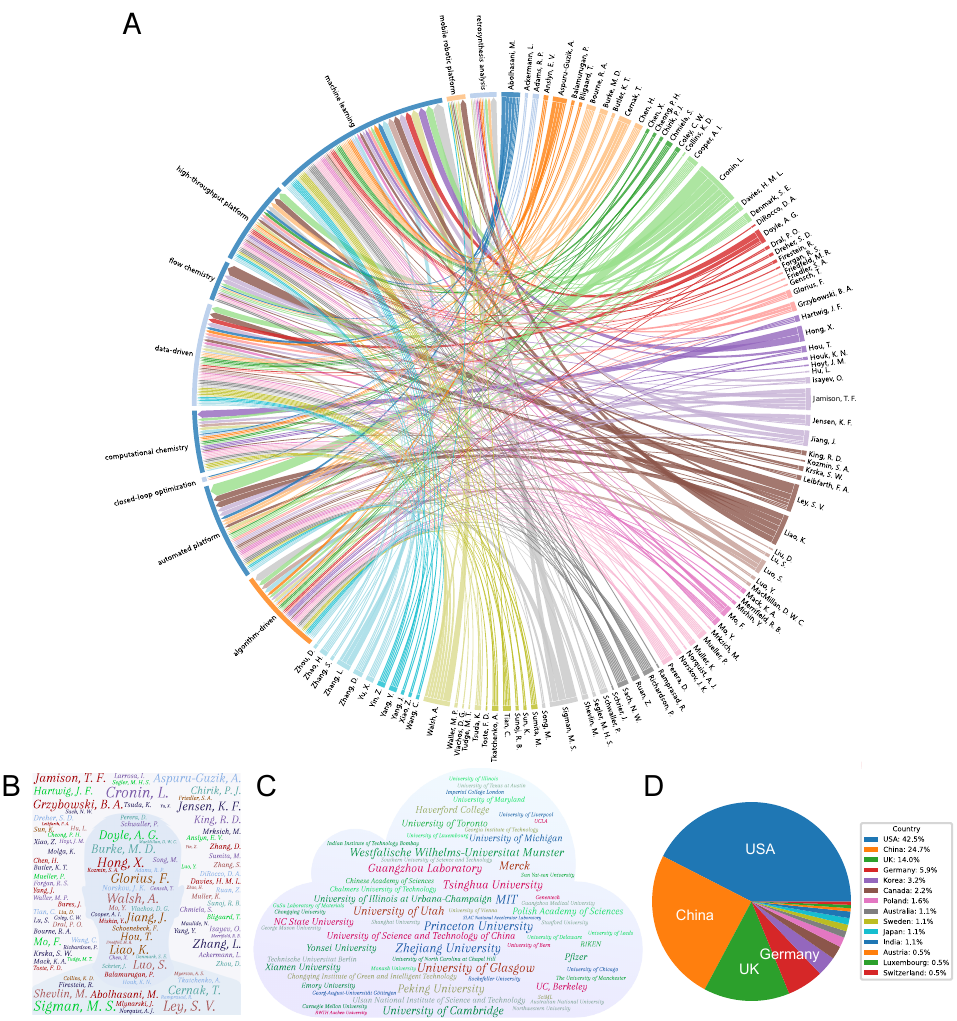}
		\caption{World-leading research group specializing in the integration of automation and artificial intelligence in organic chemistry. (A) Appraisal of the research group's diverse inputs in AI applications for organic chemistry. Visualization through (B) research groups' and (C) institute's word cloud maps, along with (D) geographical distribution.}
		\label{Figure 1}
\end{figure}

The research model of organic chemistry has changed dramatically over the years. Data, computational power, and algorithms form the triad of AI-driven scientific research \cite{6}. The advances in computational technology and the iterative refinement of algorithms in recent years have unleashed several transformative waves in science, revolutionizing traditional research methods. Chemistry, with its inherent ability to generate novel substances, is naturally well positioned to embrace this era of intelligence. Currently, scientists around the world are working together to harness the potential of AI in chemistry, giving rise to the “AI for Chemistry” movement (Figure \ref{Figure 1}).
	
As an inherently experimental discipline, chemistry, and particularly organic chemistry, demands a hands-on approach that trains the intellect as much as the physical skills. While this hands-on involvement remains integral for the foreseeable future, it also necessitates a degree of repetitive and mundane labor. At the heart of the AI for Chemistry initiative seeks to alleviate chemists from the burden of these manual tasks, enabling them to concentrate their efforts on scientifically impactful pursuits. Leveraging high-throughput robotic automation platforms, as opposed to conventional experimental models, allows for the efficient and accurate production of substantial volumes of standardized, high-quality experimental data \cite{7,8,9}. Numerous research institutions and large companies worldwide have developed a variety of such platforms with diverse functionalities. 
	
Another critical aspect of AI for chemistry is the analysis and processing of large amounts of data and the intelligent solution to scientific problems. AI models constructed using algorithms can discern patterns among multiple variables in large datasets, uncover scientific information, and reveal deep-seated logical relationships \cite{10,11,12,13}. This “data-driven” methodology greatly aids synthetic route design, chemical reaction prediction, theoretical simulation, and quantum computing, and has successfully addressed numerous scientific problems \cite{14,15,16,17}. As a result, AI-powered research methods are considered the fourth type of scientific research paradigm, following experimental science, theoretical computation, and simulation. In this article, we will review the evolution of organic chemistry research, tracing its path from labor-intensive methods to automation and finally to the integration of artificial intelligence. We will scrutinize the factors driving these paradigm shifts and evaluate their impact on efficiency, precision, and innovation in the field of organic chemistry.
	
\section{Advances in Automated Organic Synthesis Technology}
As technological capabilities continue to grow, so does the potential to automate laboratory processes. Today, automation equipment has become an integral part of laboratory operations. A key aspect of the “AI for Chemistry” initiative - the application of artificial intelligence to chemistry - is to free chemists from manual tasks, allowing them to concentrate on matters of real scientific importance. Automation technology is central to this endeavour. In biology, automation and high-throughput screening technologies were introduced into scientific research early on. In 1965, R. B. Merrifield from Rockefeller University designed the world's first automated peptide synthesizer and developed an automated solid-phase peptide synthesis technology, which led to the automation of peptide synthesis \cite{18}. In chemistry, the need for automation is even more pressing. The automation of laboratory tasks increases the overall efficiency of the experimental process by speeding up tasks, minimising waste, reducing reagent consumption and enabling higher experimental throughput \cite{19,20}. Collectively, these increased efficiencies result in reduced laboratory operating costs. The implementation of automated systems in the laboratory can relieve researchers of time-consuming and monotonous tasks, freeing up time for more specialised procedures \cite{21}. In addition, automation aims to increase the reliability and accuracy of data, as errors and variability can occur throughout the experimental process. In addition, the combination of automation and flow chemistry can improve laboratory safety by managing and safely storing hazardous materials through streamlined automation systems \cite{22}. This technology also allows for more diverse chemical synthesis route designs \cite{23}. In traditional laboratories, chemists often select reaction conditions from chemical MSDSs but avoid high-risk experiments, potentially missing optimal reaction conditions. However, with flow chemistry automation platforms, lab personnel can better avoid direct contact with these hazardous reagents and procedures \cite{24}. Numerous research institutions and large companies around the world have already developed a variety of high-throughput robotic automation platforms with different functionalities \cite{25}. In the following sections, we will discuss pipeline-based automation and robot-based automation that have had a significant impact on the field of organic synthesis.
	
\subsection{\textit{Pipeline-based Automation}}
\begin{figure}[!t]
		\centering
		\includegraphics[width=140mm, keepaspectratio]{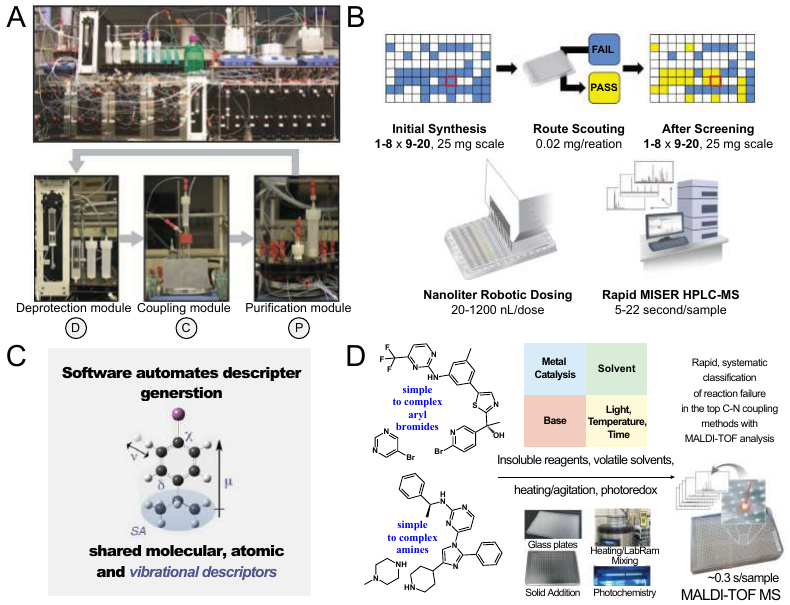}
		\caption{Revolutionary contributions related to automated and predictive chemical synthesis. (A) The creation of a fully automated synthesis platform capable of producing small molecules from 14 distinct classes. The platform's notable features include an expanded building block-based synthetic approach and capture-and-release chromatography purification systems (Adapted from ref. \cite{28}. Copyright©2015, American Association for the Advancement of Science). (B) The miniaturization of chemical experiments, facilitating large-scale synthetic processes for catalyzed cross-coupling reactions. This innovation combines nanoliter robotics with highthroughput HPLC-mass spectrometry to direct preparative-scale syntheses (Adapted from ref. \cite{29}. Copyright©2014, American Association for the Advancement of Science). (C) The use of machine learning to predict the outcomes of Buchwald-Hartwig reactions invoving isoxazoles. The system integrates with the Spartan tool, enabling calculations of molecular, atomic, and vibrational properties, which could be visually represented (Adapted from ref. \cite{31}. Copyright©2018, American Association for the Advancement of Science). (D) The automation of the Buchwald-Hartwig C-N coupling reaction via a cutting-edge synthesis platform. This process involves a workflow with MALDI-TOF MS sample analysis, optimized ionization matrix, and the application of automated liquid handling equipment (Adapted from ref. \cite{32}. Copyright©2018, American Association for the Advancement of Science)}
		\label{Figure 2}
	\end{figure}

The synthesis of small molecules often depends on highly tailored procedures specific to each target \cite{26}. If made widely available, automated processes could greatly increase the accessibility of such compounds, facilitating the exploration of their true potential \cite{27}. In 2015, M. Burke from the University of Illinois at Urbana–Champaign achieved the synthesis of 14 different classes of small molecules using a fully automated platform \cite{28}, as shown in Figure \ref{Figure 2}A. This was achieved by strategically expanding the scope of the building block-based synthetic platform to include even $Csp^3$ - rich polycyclic natural product frameworks, and devising capture-and-release chromatography purification schemes for all corresponding intermediates. With thousands of compatible building blocks now commercially available, many small molecules can be made using this platform. More generally, these findings pave the way for more universal and automated approaches to small molecule synthesis. Also in 2015, T. Cernak of the University of Michigan and S. D. Dreher of Merck explored the concept of using miniaturised chemistry experiments to guide large-scale synthetic processes of catalysed cross-coupling reactions \cite{29}, as shown in Figure \ref{Figure 2}B. The integration of high-precision nanolitre robotics, common in biochemistry laboratories, with high-throughput HPLC mass spectrometry can increase the efficiency and precision of experiments guiding preparative-scale syntheses, thereby streamlining the overall process.
	
The Buchwald-Hartwig class of Pd-catalysed C-N coupling reactions provides an essential route to the synthesis of complex drugs \cite{30}. In 2018, S. D. Dreher of Merck and A. G. Doyle of Princeton University applied machine learning to predict the outcome of the Buchwald-Hartwig reaction in the presence of isoxazoles \cite{31}, as shown in Figure \ref{Figure 2}C. Rather than focusing on the coupling of substrates with inherent heterocyclic functionality, they used the Glorius fragment additive screening technique to evaluate the effect of isoxazole fragment additives on the amination of various aryl and heteroaryl halides. Software has been developed to submit molecular, atomic and vibrational property calculations to Spartan and then extract these properties from the resulting text files for general user access, thereby avoiding the need for time-consuming analytical and computational data collection. This software can be used to predict the performance of new substrates under specific conditions or to determine the optimum conditions for new substrates. In 2018, H. Sheng, C. J. Welch and S. D. Dreher of Merck conducted a comprehensive chemical reactivity study of the Buchwald-Hartwig C-N coupling reaction using an automated synthesis platform \cite{32}, as shown in Figure \ref{Figure 2}D. This platform extended the scope of the synthesis to include matrix-assisted laser desorption/ionisation-time-of-flight mass spectrometry (MALDI-TOF MS). They optimised MALDI-TOF MS sample analysis by selecting appropriate ionisation matrices and instrument settings, and by incorporating ionisation internal standards to normalise the effect of reaction conditions. They found that $\alpha$-cyano-4-hydroxycinnamic acid (CHCA) was the optimal ionisation matrix, providing the best signal-to-noise ratio for several protonated model compounds. They also developed a workflow for MALDI analysis, including MALDI plate preparation, data acquisition and processing. In addition, the use of automated liquid handling equipment can significantly reduce sample preparation time.
	
\begin{figure}[!t]
		\centering
		\includegraphics[width=140mm, keepaspectratio]{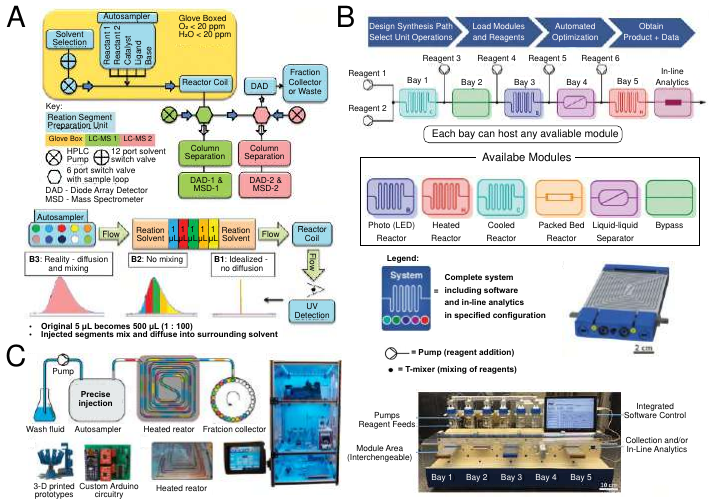}
		\caption{Flow Chemistry Based Automation Platform. (A) Pfizer's automated synthesis platform, incorporating flow chemistry, expedited nanomolar reaction screening and micromolar synthesis within a modular unit. Capable of producing data for 5,760 reactions and exceeding a rate of 1,500 reactions per day, it also yielded micromolar quantities of the desired material (Adapted from ref. \cite{37}. Copyright©2018, American Association for the Advancement of Science). (B) MIT's automated chemical synthesis platform, unveiled in 2018, merged integrated AI and robotics to minimize the professional chemist's workload. It was a versatile, plug-and-play system designed for efficient chemical synthesis (Adapted from ref. \cite{38}. Copyright©2018, American Association for the Advancement of Science). (C) The integration of flow synthesis with machine learning significantly enhanced materials discovery. This ML-guided design model identified more than 10 copolymer compositions outperforming conventional materials (Adapted from ref. \cite{40}. Copyright©2021, American Chemical Society). } 
		\label{Figure 3}
	\end{figure}

	Flow chemistry, a rapidly emerging technology in chemistry, is an innovative approach in which chemical reactions are carried out in a tube or pipe \cite{33,34,35}. Reactive components are pumped together at one end and allowed to mix and react as they flow towards the other end \cite{36}. This method offers significant advantages over traditional batch chemistry, including safer reactions, faster reactions, ease of scale-up, and greater reproducibility due to improved control over reaction parameters. In 2018, D. Perera, P. Richardson and N. W. Sach from Pfizer developed an automated synthesis platform based on flow chemistry and built from commercially available components \cite{37}, as shown in Figure \ref{Figure 3}A. This system uniquely combined rapid nanomolar reaction screening with micromolar synthesis in a modular unit. The efficacy of the system was demonstrated by investigating various reaction variables in the Suzuki-Miyaura coupling at the nanomolar scale at high temperatures. The system was able to generate data for 5760 reactions at a rate of over 1500 reactions per day and produce micromolar quantities of the desired material by multiple injections of the same segment. Optimal conditions, replicated in traditional and batch processes at the 50-200 mg scale, resulted in excellent yields. To facilitate rapid data analysis, they used Agilent ChemStation software for real-time identification of key peaks in LC-MS traces, followed by offline data optimisation using iChemExplorer software and visualisation using Spotfire.
	
In 2018, T. F. Jamison and K. F. Jensen from MIT developed a plug-and-play automated chemical synthesis platform based on flow chemistry \cite{38}, as shown in Figure \ref{Figure 3}B. The aim of this platform was to reduce the workload of professional chemists and enhence their capacity for perform targeted process synthesis. In 2021, O. Isayev from Carnegie Mellon University and F. A. Leibfarth from the University of North Carolina at Chapel Hill developed a computational materials discovery approach that combined automated flow synthesis with machine learning to develop advanced copolymers for $^{19}$F MRI imaging agents \cite{40}, as shown in Figure \ref{Figure 3}C. They created a software-controlled continuous polymer synthesis platform for iterative experimental and computational cycles, exploring 0.9\% of the total composition space and synthesising 397 different copolymer compositions. The ML-driven design criteria identified more than 10 copolymer compositions that surpassed the performance of existing materials. Despite these advances, synthetic routes still require human input and process development. However, the combination of automation and machine learning techniques has the potential to revolutionise the field of chemical synthesis, opening up avenues for the discovery of new and innovative compounds with significant scientific and industrial applications.
	
\begin{figure}[!t]
		\centering
		\includegraphics[width=140mm, keepaspectratio]{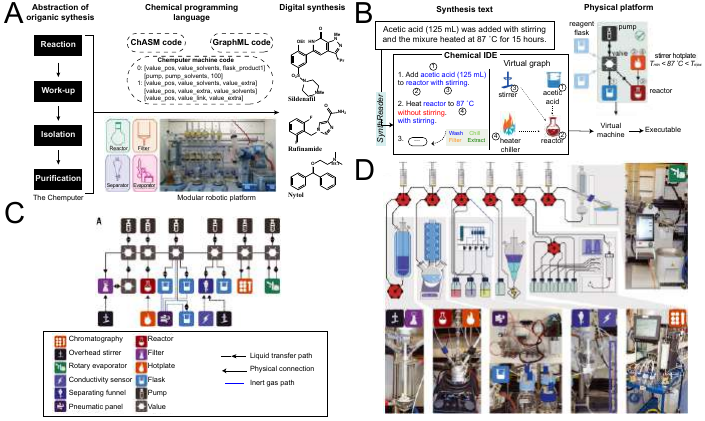}
		\caption{Illustrations depict various innovations in the field of automated chemical synthesis and robotics. (A) A standalone plastic reactor facility has been developed for producing fine chemicals and pharmaceuticals (Adapted from ref. \cite{57}. Copyright©2019, American Association for the Advancement of Science). This facility was designed using a chemistry-to-computer-automated design (ChemCAD) approach, which converts conventional lab-scale synthesis into a universally compatible digital code. (B) A scalable chemistry execution architecture has been devised that can autonomously populate itself by scanning academic literature, leading to a universally applicable autonomous workflow (Adapted from ref. \cite{60}. Copyright©2020, American Association for the Advancement of Science). The resultant robot-generated code is human-readable and doesn't require any programming expertise to correct. It also transcends specific hardware due to the standardization. (C) , (D) ChemPU is an automated chemical reaction database that houses codes for 100 distinct reactions prevalent in modern organic synthesis (Adapted from ref. \cite{61}. Copyright©2022, American Association for the Advancement of Science). It features a variety of reactions such as transition metal-catalyzed coupling, heterocycle formation, functional group interconversions, and multicomponent reactions. The codes, known as ÇDL, are stored in a database that allows version control, validation, collaboration, and data mining.}
		\label{Figure 6}
	\end{figure}

Chemical reaction discovery is inherently unpredictable and time-consuming. Reactivity prediction is an attractive alternative, however, related methods such as computational reaction design are still in their infancy. Reaction prediction based on advanced quantum chemical methods is complex, even for simple molecules. In 2018, L. Cronin from the University of Glasgow proposed a method for producing fine chemicals and pharmaceuticals in a self-contained plastic reactor device \cite{57}, as shown in Figure \ref{Figure 6}A. The device was designed and built using a chemistry-to-computer-automated design (ChemCAD) approach, which translates traditional lab-scale synthesis into platform-independent digital code. This code guides the production of 3D printed devices that contain the entire synthesis route internally. With the production of the GABA receptor agonist (±)-baclofen, they established a concept for localised drug production outside of specialist facilities. In 2018, they demonstrated the concept by guiding an automated system to synthesise three drugs (diphenhydramine hydrochloride, rufinamide and sildenafil) without human intervention \cite{58}. In 2018, L. Cronin developed an organic synthesis robot capable of performing and analysing chemical reactions faster than manual methods by efficiently navigating the chemical reaction space by predicting the reactivity of potential reagent combinations after a limited number of experiments \cite{59}. In 2020, L. Cronin developed a scalable chemistry execution architecture that can be automatically populated by reading the literature, resulting in a generic autonomous workflow \cite{60}, as shown in Figure \ref{Figure 6}B. The robot-synthesised code is correctable in natural language without any programming knowledge and is hardware independent due to standards. This chemical code can then be combined with a graph describing the hardware module and compiled into platform-specific low-level robot instructions for execution. To validate the method, the automated synthesis of 12 compounds was performed, including the analgesic lidocaine, the Dess-Martin periodinium oxidation reagent and the fluorinating agent AlkylFluor. In 2022, L. Cronin developed ChemPU, an automated chemical reaction database containing 100 molecules representing the range of reactions found in modern organic synthesis \cite{61}, as shown in Figure \ref{Figure 6}C and \ref{Figure 6}D. These reactions include transition metal-catalysed coupling reactions, heterocycle formation, functional group interconversions and multicomponent reactions. The chemical reaction codes, or $\chi$DL, of the reactions were stored in a database for version control, validation, collaboration and data mining. In 2022, they also used the system to develop an autonomous portable platform for general chemical synthesis and demonstrated the system by synthesising five small organic molecules, four oligopeptides, and four oligonucleotides with good yield and purity, performing a total of 24,936 base steps in 329 hours of platform runtime \cite{62}.
	
\begin{figure}[!t]
		\centering
		\includegraphics[width=140mm, keepaspectratio]{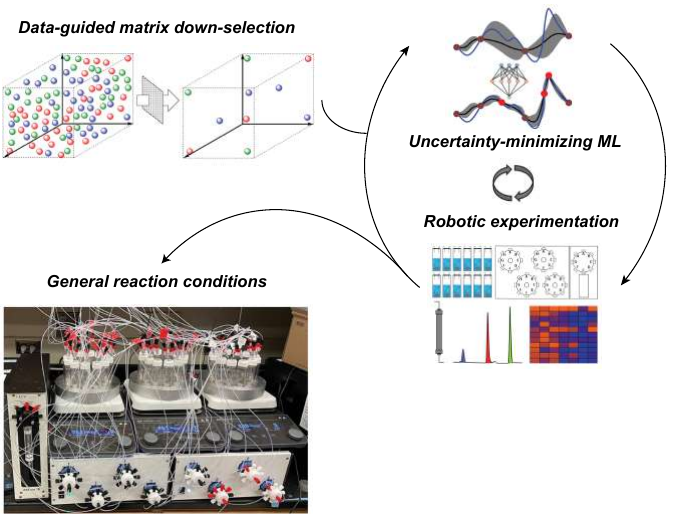}
		\caption{Autonomization of Synthetic Chemistry. The closed-loop optimization process integrates data-driven and machine learning strategies to build effective models. This subsequently guides robotic systems in performing automated experiments with reduced iterative costs. Optimization of heteroaryl Suzuki-Miyaura coupling reactions, with the validity confirmed under universal experimental conditions (Adapted from ref. \cite{68}. Copyright©2022, American Association for the Advancement of Science).}
		\label{Figure 8}
	\end{figure}

	The promise of fully automated chemical synthesis is broad and on-demand access to small molecules, a prospect that could revolutionise the field \cite{65}. However, the range of reactions that can be performed autonomously remains limited. Automated stereospecific construction of $Csp^3$-C bonds will broaden our access to a variety of important functional organic molecules \cite{66}. Historically, methyliminodiacetic acid (MIDA) boronate has been used to coordinate the formation of $Csp^2$-$Csp^2$ bonds and has become a critical building block for the automated synthesis of many small molecules. However, this is incompatible with the reactions that form stereospecific $Csp^3$-$Csp^2$ and $Csp^3$-$Csp^3$ bonds \cite{67}. In 2022,  D. J. Blair from the University of Illinois at Urbana-Champaign and M. D. Burke introduced a new tetramethyl-N-methyliminodiacetic acid (TIDA) boronic acid ester \cite{66}. This new compound demonstrated stability under reaction conditions that would normally be considered challenging. Analysis revealed that the redistribution of electron density increased the covalency of the N-B bond, thereby reducing the degree of hydrolysis. Complementary steric shielding of the carbonyl À face reduces its reactivity towards nucleophiles. TIDA boronate retains the unique properties of the iminodiacetic acid cage, which is critical for large-scale automated synthesis. This allows the $Csp^3$ borate building block to be used in automated synthesis, including the preparation of natural products by automating stereospecific $Csp^3$-$Csp^2$ and $Csp^3$-$Csp^3$ bond formation. These discoveries will facilitate the automated assembly of more complex small molecules containing large amounts of $Csp^3$ . In 2022, B. A. Grzybowski from Ulsan National Institute of Science and Technology and M. D. Burke developed a workflow concept for closed-loop optimization \cite{68}, as shown in Figure \ref{Figure 8}. Closed-loop optimization involves building a model using data-driven and machine learning methods, and then directing the robot to perform automated experiments with the lowest trial and error cost according to the model guidance. This culminated in the optimized heteroaryl Suzuki-Miyaura coupling reaction, validated by experimental conditions of universality.
	
	\subsection{\textit{Robot-Based Automation}}
	
	\noindent The advent of robotic chemistry experiments signalled the inception of the AI-powered laboratory. The aspiration of automated chemists is to establish a fully automated workflow. Nonetheless, the majority of current mechanised chemistry platforms are limited to partial automation, posing significant challenges for comprehensive scientific research. A notable advancement towards achieving a fully automated chemistry workflow was the introduction of a mobile robot by A. I. Cooper from the University of Liverpool in 2020 \cite{56}. Powered by a batch Bayesian search algorithm, the robot autonomously conducted 688 experiments within a 10-variable experimental space over a span of eight days.
	
\begin{figure}[!t]
		\centering
		\includegraphics[width=140mm, keepaspectratio]{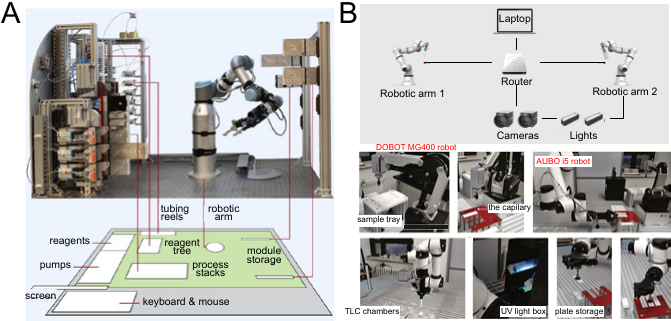}
		\caption{The figure depicts the robot-based automation platform, which powered by scientific data intelligence, excels at handling routine tasks common in chemical research. (A) In 2019, the ASKCOS software program was introduced, integrating artificial intelligence for strategic synthetic planning and robotics for efficient execution  (Adapted from ref. \cite{39}. Copyright©2019, American Association for the Advancement of Science). (B) High-throughput Thin Layer Chromatography (TLC) Robotic Platform (Adapted from ref. \cite{43}. Copyright©2022, Elsevier)}
		\label{Figure 13}
	\end{figure}

	Standardizing processes, bolstering the reliability of experiments, and curbing expenses are pivotal reasons for automating chemical experiments. This is particularly significant as it enables the generation of comprehensive and resilient datasets, a crucial element for ensuing statistical assessments. A revolutionary strategy, merging automation and artificial intelligence (AI) techniques, has been successfully implemented across diverse facets of chemical science, from refining reaction variables to unraveling fundamental mechanisms. In 2019, T. F. Jamison and K. F. Jensen from MIT augmented the robotic platform by incorporating artificial intelligence for synthetic planning and robotics for execution \cite{39}, as depicted in Figure \ref{Figure 13}A. The system utilized computer-aided synthetic programming (CASP) to determine synthetic routes via a root-parallel Monte Carlo tree search, while a robotic arm carried out the reactions to synthesize compounds as specified in the chemical recipe file. This platform successfully automated the synthesis of 15 medically relevant small molecules, employing different synthetic routes based on the target molecule's complexity. Concurrently, they developed a novel software program, ASKCOS, which leverages the collected data to predict organic chemical reaction outcomes.
	
	In 2022, F. Mo and D. Zhang from Peking University further contributed to the development of robotic platforms by constructing a custom desktop robot system for high-throughput thin-layer chromatography (TLC) data collection \cite{43,44}, as illustrated in Figure \ref{Figure 13}B. Executing a TLC analysis requires a series of methodical steps: first the analyte is dissolved, then spotted, followed by the development of the TLC plate, and concluding with the measurement and calculation of each analyte's $R_f$ value. Central to this process are two synchronized robots, designed to carry out complex tasks with the utmost precision and safety, reminiscent of the dexterity of human limbs. The apparatus also incorporates a pair of cameras, a duo of light sources, a networking device, and a computing machine. The lesser of the two robots, the DOBOT MG400, is tasked with the preparation and application of the TLC samples onto the designated plates. This is followed by the AUBO i5, its larger counterpart, which carefully holds and positions the TLC plate into the designated chamber for the development phase. Once this phase is complete, the plate is relocated to a receptacle for visualization and subsequent imaging under ultraviolet illumination. The AUBO i5 then secures a fresh plate from its storage location and places it onto the designated stand, thereby resetting the system for subsequent iterations. To enhance operational efficiency, the setup includes six chambers, each containing a unique elution solvent. A specially designed Python script oversees the entire operation, managing the robots, cameras, and illumination sources. This state-of-the-art robotic system for TLC experiments is pivotal for enabling the standardized, high-throughput acquisition of $R_f$ values
	
	\subsection{\textit{Selection of the Appropriate Automation Approach}}
	\noindent Automated systems in organic chemistry encompass a spectrum of technologies designed to streamline and optimize chemical processes, from the initial mixing of reactants to the final purification and isolation of products \cite{104}. Two prevalent approaches are pipeline-based automation and robot-based automation. Pipeline-based automation involves a series of automated processes and machines, organized sequentially (in a pipeline), to facilitate different stages of a chemical reaction or process \cite{105}. Conversely, robot-based automation leverages robots to execute tasks traditionally performed by human operators, ranging from simple tasks, like weighing and mixing reactants, to more complex ones, like monitoring reactions, purifying products, and analyzing results \cite{39,43,44}. Robots can be programmed to execute tasks sequentially or in parallel and can be equipped with various tools and sensors to perform diverse operations. By assessing these factors, researchers and practitioners can make informed decisions on the most suitable automation approach for their specific needs, as outlined below:
	\begin{itemize}
		
		\item[•] Efficiency: Pipeline-based automation is generally more efficient for large-scale production as it can manage large volumes of samples or reactions in a continuous flow manner. In contrast, robot-based automation is more efficient for tasks necessitating high levels of precision and accuracy.
	
		\item[•] Flexibility: Robot-based automation offers greater flexibility and adaptability to changing requirements or new reactions, as robots can be reprogrammed to perform different tasks. Conversely, pipeline-based automation is less flexible due to its fixed setup, with each stage of the process carried out in a specific order.
		
		\item[•] Safety: Both pipeline-based and robot-based automation enhance safety by minimizing personnel exposure to hazardous chemicals and reaction conditions. However, robot-based automation may confer additional safety benefits, as robots can handle hazardous chemicals and operate under dangerous conditions without exposing human operators to risk.
		
		\item[•] Cost: Both automation approaches necessitate a significant initial investment. The cost of pipeline-based automation includes setting up the pipeline and associated machinery, while the cost of robot-based automation includes purchasing robots and associated equipment. The overall cost will vary depending on the reaction or process complexity and production scale.
		
		\item[•] Maintenance: Regular maintenance is essential for both automation approaches to ensure proper functioning. Maintenance of pipeline-based automation involves ensuring all pipeline parts are functioning correctly, whereas maintenance of robot-based automation entails ensuring the robot and its associated equipment are functioning correctly.
	\end{itemize}
	
	Ultimately, the choice between pipeline-based and robot-based automation will hinge on various factors, such as production scale, reaction or process complexity, the need for flexibility, and the available budget. Often, a combination of both approaches may be utilized to optimize the overall process. For instance, a robot could be employed to prepare and load samples into a pipeline-based system for continuous flow reactions \cite{103}.
	\section{AI-Facilitated Organic Chemistry Research}
	
AI for Chemistry is an emerging discipline that focuses on using large-scale data analysis and intelligent decision making to address scientific challenges. Through algorithmic processes, AI models can identify patterns in large data sets, extract scientific insights and discover complex logical relationships. Since 1969, E. J. Corey has pioneered the integration of computational and graphical communication technologies into organic synthesis pathways, thereby charting a new course for the design of complex organic synthesis pathways \cite{41}. The concept of explicable AI and AI-driven hypothesis generation is a relatively recent development in chemistry. Recently, A. Aspuru-Guzik and his team from the University of Toronto developed a machine learning algorithm capable of extracting human-understandable insights from large datasets in chemistry and physics \cite{42}. This algorithm not only confirmed some well-known “rules of thumb” about the solubility and energy levels of organic molecules, but also revealed new principles. In subsequent sections, we delve into research in organic chemistry facilitated by artificial intelligence. We will explore AI applications in this domain, including retrosynthetic analysis, predicting molecular properties, and forecasting the reactivity of chemical reactions.
	
	\subsection{\textit{AI-Powered Retrosynthetic Analysis Models}}
	Retrosynthetic analysis stands as a pivotal method in the systematic planning for the synthesis of small organic molecules, having grown significantly due to recent advancements in technology \cite{52,53}. A monumental achievement in this field is credited to B. A. Grzybowski and his team, who birthed a revolutionary retrosynthetic analysis software named Chematica \cite{106}. This sophisticated software is founded on a vast database comprising over 7 million organic molecules, meticulously interconnected through an equal number of organic reactions to form a comprehensive network. A monumental endeavor, the team input over 50,000 organic reaction rules manually, instructing Chematica on the potential alterations each small molecule might undergo during reactions. It should be noted that the exact figures might be subjected to updates. In various trials, Chematica curated synthesis routes that not only diverged significantly from those previously reported by chemists but also showcased fewer steps and higher yields, thereby reducing time and costs markedly. This groundbreaking tool transitioned into a new phase in 2017 when it was acquired by Merck, undergoing a rebranding to become Synthia for commercial service.
	
Taking an enormous leap forward, the development transitioned from a manual rule input system to an autonomous learning entity. This evolution was actualized in 2018 when researchers M.S.H. Segler from Westfälische Wilhelms-Universität Münster and M.P. Waller from Shanghai University augmented it with a Monte Carlo tree search (MCTS) integrated with an extended policy network to steer the search process, coupled with a filter network to prioritize the most promising retrosynthetic pathways \cite{54}. They trained these deep neural networks on an extensive set of published organic chemistry reactions. The system efficiently solved almost twice the number of molecules and was 30 times faster than traditional computational search methods, which rely on extracted rules and custom-designed heuristic algorithms. They also performed BFS at a cost determined by a policy network, called ``neural BFS''. All algorithms used the same set of automatically extracted transformations. Double-blind AB experiments confirmed the superiority of 3N-MCTS over traditional methods, with organic chemists clearly favouring the former. Machine-generated retrosynthetic routes should be considered on a par with reported molecular routes of practical interest to organic chemists. 
	
	In 2021, P. Schwaller from the University of Bern demonstrated that transformer neural networks can learn atomic mapping information between reactants and products without supervision or manual labelling \cite{55}. They developed a chemically neutral, attention-guided reaction mapper called RXNMapper, which uses the attention weights of transformers to extract coherent chemical grammars from unannotated reaction sets. RXNMapper excels at carbon skeleton rearrangements and complex reactions, which often require human understanding of the reaction mechanism for accurate atom maps. The method achieved 99.4\% correct complete atom mappings on a test set of 49,000 highly unbalanced patent reactions. This approach outperforms others in both accuracy and speed, even for reactions with non-trivial atomic mappings and chemical complexity. It effectively bridges the gap between data-driven and rule-based strategies in various chemical reaction tasks.

	\subsection{\textit{Prediction of Molecular Properties}}
	\noindent Predicting the properties of compounds is critical to numerous applications such as drug design and materials science. The emergence of computational methods as potent instruments has enabled precise predictions of molecular attributes, thereby facilitating the strategic development of compounds across different sectors. This subsection delves into data derived from theoretical calculations and high-throughput experiments, enhancements in computational capacities for modeling polarization and determining molecular characteristics, the creation of automated systems for thin-layer chromatography (TLC) and high-performance liquid chromatography (HPLC), and the incorporation of machine learning methodologies for analyzing and predicting compound attributes and separations.
	
	The advancement of computational capabilities has opened new avenues for modeling polarization. P. L. A. Popelier, in 2014, showcased that kriging, a machine learning method, can precisely interpret the electron density response of a topological atom to changes in the positions of nearby atoms \cite{100}. An innovative approach of altering training set sizes was employed, drastically reducing training times while retaining accuracy. This approach was implemented for each amino acid, modifying their geometries through normal modes of vibration across all local energy minima on the Ramachandran map. The modified geometries served as the training data for the kriging models. The validity of the kriging models was assessed through a meticulous comparison of the predicted and actual total electrostatic energies for previously untested geometries. This approach was substantiated using aromatic amino acids such as histidine, phenylalanine, tryptophan, and tyrosine as test cases. 
	
	The ability to accurately and efficiently predict molecular properties across the chemical compound space is pivotal for the rational design of compounds in the chemical and pharmaceutical industries. To address this, A. Tkatchenko, in 2015, developed and implemented a series of efficient empirical methods for estimating atomization and total energies of molecules \cite{101}. The methods ranged from a simple sum over atoms, the addition of bond energies, pairwise interatomic force fields, and extended to advanced machine learning approaches capable of describing collective interactions among multiple atoms or bonds. Even simple pairwise force fields exhibited prediction accuracy comparable to benchmark energies computed using density functional theory with hybrid exchange-correlation functions for equilibrium molecular geometries. This remarkable accuracy was achieved using a vectorized representation of molecules, the Bag of Bonds model, which exhibited strong nonlocality in chemical space. Additionally, this representation facilitated the accurate prediction of electronic properties of molecules, such as polarizability and molecular frontier orbital energies. 
	
	Thin layer chromatography (TLC) is an important tool for monitoring chemical reactions and discerning the ability of different substances to separate. In 2022, F. Mo and D. Zhang of Peking University combined TLC with automation and artificial intelligence to make an excellent work for the prediction of TLC \cite{43}, as shown in Figure \ref{Figure 13}A. This remarkable innovation aims to facilitate the collection of large volumes of TLC data using a specialized desktop robotic system to meticulously handle the complex steps involved in TLC analysis. The platform excels at performing large amounts of standardized $R_f$ value assessments, resulting in large amounts of standardized TLC data. The next step is to perform feature engineering on the raw data. The process of feature engineering allows for the translation of compound structures, additional physical properties, and varying factors such as elution solvents into a vector space, thereby streamlining the data cleaning process. Following this, a series of machine learning methodologies are applied to perform regression analysis on the engineered data. The predictive accuracies of multiple machine learning methods, including Bayesian regression, Random Forest (RF), Light Gradient Boosting Machine (LightGBM or LGB), Extreme Gradient Boosting (XGBoost or XGB), and Artificial Neural Network (ANN), were carefully evaluated. Building on this foundation, an ensemble approach is suggested to construct a superior model, thereby enhancing the accuracy further. A model boasting the highest accuracy, with an $R^2$ value of 0.951, is achieved by employing a simple weighted average of the aforementioned methods. The commendable results also underscore the precision and reliability of the dataset acquired through the automated platform, as machine learning techniques impose strict demands for accurate, independent, and identically distributed data. This research offers invaluable insights into the selection of optimal purification parameters and presents organic chemists with a pragmatic artificial intelligence tool for chromatography applications.
	\begin{figure}[!t]
		\centering
		\includegraphics[width=140mm, keepaspectratio]{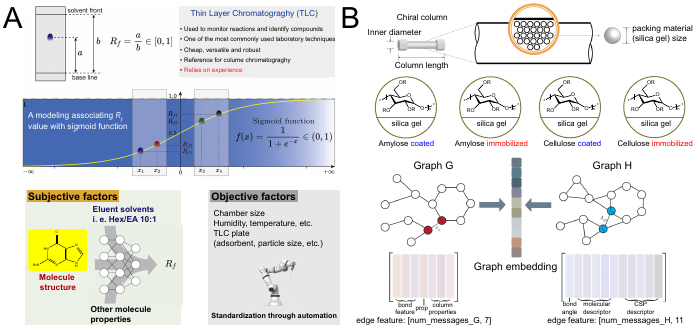}
		\caption{AI-Driven Prediction of Compound Separation (A) High-throughput Thin Layer Chromatography (TLC) Robotic Platform: This sophisticated platform is designed for large-scale standardization and measurement of $R_f$ values, subsequently generating substantial volumes of normalized TLC data. This data is then utilized for regression analysis through machine learning techniques  (Adapted from ref. \cite{43}. Copyright©2022, Elsevier). (B) Predicting Chromatographic Enantiomer Separation: By leveraging a literature-based dataset (CMRT dataset) of retention times for chiral molecules in HPLC, establishing a correlation between molecular structure and retention time. This is further enhanced by employing the Quantile Geometry Augmented Graph Neural Network (QGeoGNN) for achieving optimal prediction accuracy (Adapted from ref. \cite{45}. Copyright©2022, Springer Nature).}
		\label{Figure 4}
	\end{figure}
	
	High performance liquid chromatography (HPLC) is a technique used to separate, identify and quantify each component in a sample mixture based on differences in adsorption capacity with an adsorbent material. In 2023, F. Mo and D. Zhang introduced machine learning methods to chromatographic enantiomer separation \cite{45}, as shown in Figure \ref{Figure 4}B. They developed a literature dataset (CMRT dataset) of retention times of chiral molecules in HPLC to address the challenge of data acquisition. Using this dataset, they introduced a Quantile Geometry Augmented Graph Neural Network (QGeoGNN) to detect the relationship between molecular structure and retention time, demonstrating promising predictive power for enantiomers. The data set includes different HPLC columns with two substrates and seven substituents. Their study achieved multi-column prediction by incorporating chromatographic domain knowledge into a machine learning model, paving the way for predicting chromatographic enantiomeric separations by calculating separation probabilities. Their work highlights the potential application of machine learning techniques in experimental settings, ultimately accelerating the pace of scientific discovery.
	
	\subsection{\textit{Prediction of Reactivity of Chemical Reactions}}
	
	In the realm of chemical reaction predictions, Multivariate Linear Regression (MLR) has stood out as a fundamental statistical instrument. While traditional linear regression hinges on one predictor variable, MLR takes into account multiple predictors, painting a holistic picture of their collective impact on the outcome.
	
\begin{figure}[!t]
		\centering
		\includegraphics[width=140mm, keepaspectratio]{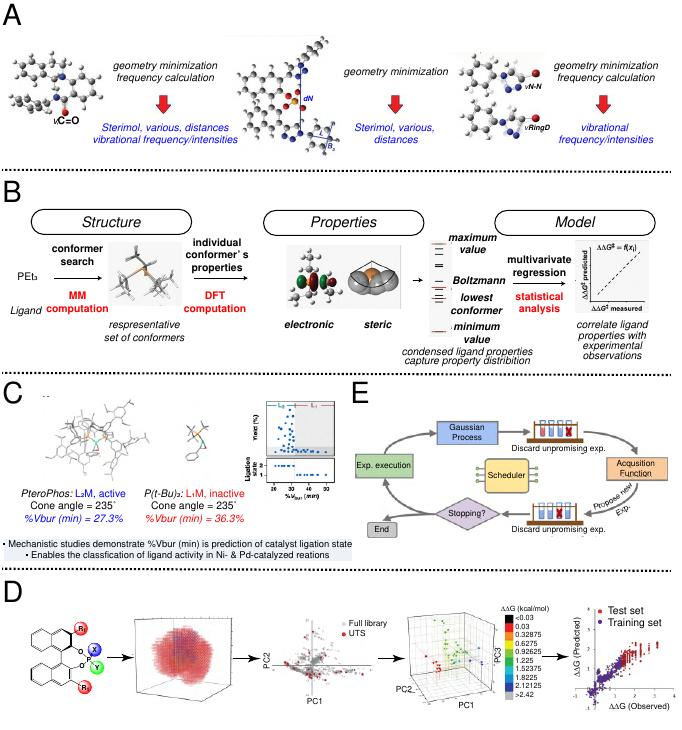}
		\caption{AI-Driven Catalyst Design and Optimization. (A) Data-intensive approach to Chiral Anionic Catalysis: Introduced by F. D. Toste and M. S. Sigman in 2015, this approach entailed a comprehensive analysis of various catalysts and substrates, revealing the key mechanistic attributes of the system and insights into enantioselectivity (Adapted from ref. \cite{46}. Copyright©2015, American Association for the Advancement of Science). (B) Stereodivergent Pd-catalyzed Cross-Coupling Reactions: M. S. Sigman and M. R. Biscoe employed molecular modeling and parameterization in conjunction with experimental validation to elucidate the factors governing stereoselectivity in 2018 (Adapted from ref. \cite{47}. Copyright©2018, American Association for the Advancement of Science). (C) Monodentate Phosphine Ligands Classification: In 2021, M. S. Sigman and A. G. Doyle presented a methodology to categorize monodentate phosphine ligands based on their connectivity status and reactivity, enabling a more nuanced understanding of reaction outcomes in cross-coupling catalysis(Adapted from ref. \cite{49}. Copyright©2021, American Association for the Advancement of Science). (D) Computationally Guided Catalyst Selection: S. E. Denmark demonstrated in 2019 how machine learning and cheminformatics could revolutionize the selection of chiral catalysts (Adapted from ref. \cite{50}. Copyright©2019, American Association for the Advancement of Science). (E) Automatic Reactive Optimization and Parallel Scheduling (AROPS): Introduced by Y. Mo in 2023, AROPS effectively integrates optimization algorithms with module scheduling, resulting in a more efficient experimental process (Adapted from ref. \cite{51}. Copyright©2023, American Chemical Society).}
		\label{Figure 5}
	\end{figure}

	The pivotal role of MLR in organic chemistry research can be attributed to M. S. Sigman's seminal research, particularly in forecasting specific reaction outcomes—like discerning the enantioselectivity associated with diverse catalyst structures. In a pioneering series of investigations, Sigman collaborated with esteemed peers to spearhead advancements in chemical research. A highlight of his scientific journey was in 2015 when Sigman joined forces with F. D. Toste from the University of Utah. Their data-centric approach illuminated the intricacies of asymmetric induction in chiral anionic catalysis, taking a deep dive into the interaction of various catalysts and substrates \cite{46}, as shown in Figure \ref{Figure 5}A. In 2018, Sigman worked alongside M. R. Biscoe from the City College of New York to delve into stereodivergent Pd-catalysed cross-coupling reactions. They employed a blend of molecular modeling, parameterization, and experimental validation to refine reaction conditions and identify critical factors influencing stereoselectivity \cite{47}, which is demonstrated in Figure \ref{Figure 5}B. Building upon this, in 2019, Sigman unveiled an innovative modeling method for enantioselectivity. Through crafting mechanism-specific correlations, the study revealed patterns in response and supplied a mechanistic rationale for certain responses \cite{48}. Futhermore, in 2021, Sigman together with A. G. Doyle from UCLA, formulated an intricate classification system highlighting the interplay and reactivity trends of monodentate phosphine ligands. This insightful approach illuminates the effect of connectivity on reaction results and presents a side-by-side analysis of nickel and palladium catalysis. Such a classification mechanism can fast-track mechanistic examinations of akin organometallic reactions, aiding in the design of reactions by predicting the behavior of mono- and bi-bound phosphines. This methodology stands as a pivotal tool for grasping structure-reactivity associations in catalysis \cite{49}, as depicted in Figure \ref{Figure 5}C.
	
	As the development of multiple linear regression (MLR) progresses, the scientific community is increasingly attracted to nonlinear machine learning (ML) methodologies. Notable techniques such as Random Forests and Neural Networks have gained popularity due to their adaptability. Often, these techniques outperform MLR, especially in situations characterized by complex features and large datasets. However, these advanced methods come with their own set of challenges, including interpretability issues, susceptibility to overfitting, and high computational demands. Historically, the design of asymmetric reaction catalysts relied on empirical methods, which required researchers to intuitively interpret structural nuances to improve selectivity. Fast forward to today, and we see machine learning and cheminformatics integrating seamlessly with traditional methods, enabling quick trend identification in extensive datasets. A significant shift occurred in 2019, thanks to the pioneering work of S. E. Denmark. Denmark's computational framework cleverly combined advanced cheminformatics techniques, representing a dramatic departure from conventional methods \cite{50}, as illustrated in Figure \ref{Figure 5}D. At the heart of his innovation were the invariant molecular descriptors, valued for their stability against structural modifications of catalysts. These descriptors formed the foundation of a meticulously curated training database, highlighting the steric and electronic intricacies of catalysts. Leveraging this wealth of data, machine learning tools, specifically support vector machines and deep feed-forward neural networks, were fine-tuned. Their effectiveness was demonstrated by their unparalleled predictive abilities, particularly in chiral phosphoric acid-mediated thiol additions to N-imides. Moreover, the Universal Training Set (UTS) denotes a strategically chosen representative subset of catalysts that is agnostic to the reaction or mechanism, making it apt for optimizing any reaction catalyzed by a specific scaffold. The Kennard-Stone algorithm was employed to select the UTS of phosphoric acid catalysts, ensuring the sampling of catalysts from uniform regions of the feature space. This approach boosts confidence in predictions made during method development as they remain within the feature space defined by the initial training set. Hence, the UTS is a potent tool for optimizing reactions catalyzed by this phosphoric acid catalyst, and we anticipate a growing number of reactions optimized using this method in the future.
	
	While innovations in automation and algorithms have enriched the field, a synchronicity deficit between platforms and optimization algorithms remained. This hurdle was adeptly surmounted in 2023 by Y. Mo of Zhejiang University with the debut of the Automatic Reactive Optimization and Parallel Scheduling (AROPS) platform \cite{51}, as shown in Figure \ref{Figure 5}E. The linchpin of AROPS is its bespoke Bayesian optimizer, tailored for managing the nuances of multi-reactor/analyzer reactive optimization. Boasting three adaptive scheduling modes, it also possesses a unique strategy grounded in the Probability of Improvement (PI), enabling the early cessation of unpromising experiments, thus refining resource deployment in simultaneous optimization tasks. Validated against cornerstone reactions in organic synthesis, AROPS epitomizes the aspirations of the modern chemist, harmonizing resourcefulness with innovation.
	
\begin{figure}[!t]
		\centering
		\includegraphics[width=140mm, keepaspectratio]{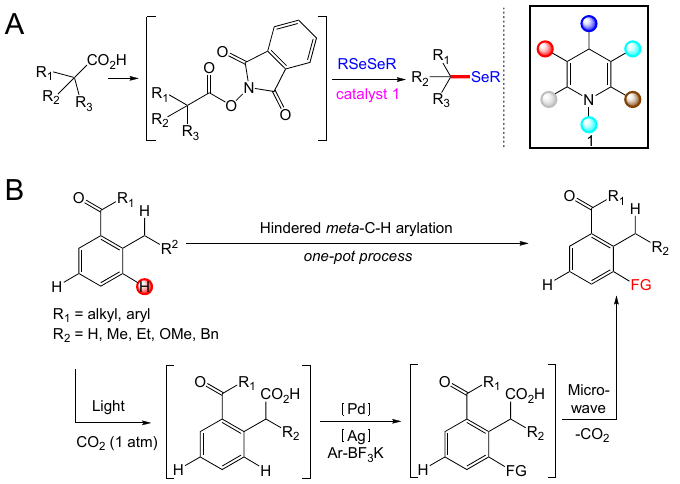}
		\caption{Enhancing Chemical Synthesis via the Integration of High-Throughput Experimentation and Artificial Intelligence. (A) The intersection of automated high-throughput experimentation (HTE) and deep learning (DL) technologies to boost efficiency in chemical synthesis (Adapted from ref. \cite{76}. Copyright©2022, Elsevier). (B) Exploring the use of 1,4-dihydropyridine (DHP) derivatives as catalysts in the decarboxylation selenization reaction (Adapted from ref. \cite{77}. Copyright©2023, ROYAL SOCIETY OF CHEMISTRY). Broad substrate versatility of the reaction is significantly enhanced by integrating HTE and AI.}
		\label{Figure 10}
	\end{figure}

	The increasing demand for accuracy and efficiency in scientific research has spurred the development of fully automated workflows in chemistry, significantly changing traditional laboratory practices \cite{73,74,75}. These automated systems increase efficiency by allowing round-the-clock experimentation and minimising human error, thereby improving the accuracy of results. In this context, the historically difficult task of selectively functionalising hindered aromatic meta-C-H bonds saw progress in 2022. K. Liao from Guangzhou Laboratory and Y. Yang from Sun Yat-sen University successfully addressed this challenge by harnessing the power of automation-driven high-throughput experimentation (HTE) and deep learning (DL) \cite{76}, as shown in Figure \ref{Figure 10}A. They introduced a novel method for functionalising these obstructed C-H bonds using carbon dioxide as an unobtrusive director in a one-pot, three-step process. This process has been specifically designed for the selective arylation of o-alkylaryl ketones at the obstructed meta position. The method includes light-induced C-H carboxylation, carboxyl-directed Pd-catalysed C-H functionalisation and microwave-assisted decarboxylation. Using HTE and DL, they studied a wide range of substrates, including over a thousand (1,032) reactions, and developed a DL-driven reaction yield prediction model called CMPRY. This model was evaluated in two different tests using unseen o-alkylaryl ketones and/or potassium aryltrifluoroborates, and demonstrated remarkable predictive performance with mean absolute yield errors of only 6.6\% and 8.4\%. Such results underline its potential for real-world synthetic applications. Meanwhile, in 2023, K. Liao and H. Chen from the Guangzhou laboratory made a breakthrough by innovatively using 1,4-dihydropyridine (DHP) derivatives - typically known for their biological relevance - as catalysts in a decarboxylative selenation reaction \cite{77}, as shown in Figure \ref{Figure 10}B. This reaction demonstrated extensive substrate versatility, which was enhanced by high throughput experimentation (HTE) and artificial intelligence. The AI-driven model demonstrated the ability to identify key structural features and accurately predict previously unseen reactions. With a correlation coefficient of 0.89, a root mean square error of 9.0\% and a mean absolute error of 6.3\%, this model demonstrated remarkable performance. This work not only expanded the catalytic applications of DHP derivatives but also highlighted the effectiveness of integrating HTE and AI to advance the development of chemical synthesis.
	
\begin{figure}[!t]
		\centering
		\includegraphics[width=140mm, keepaspectratio]{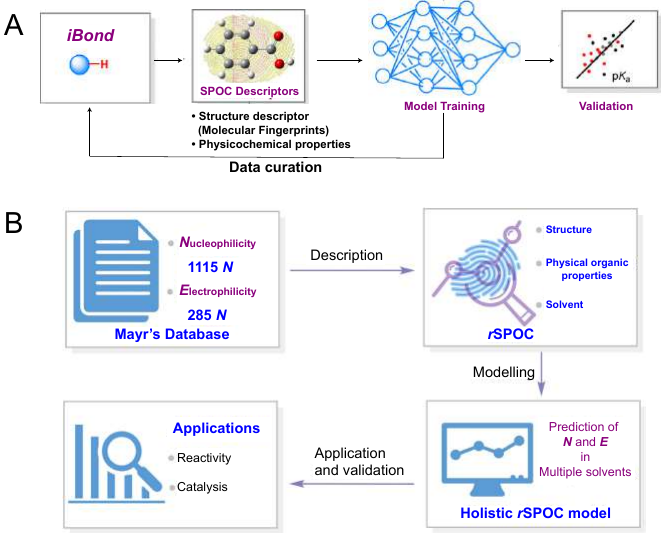}
		\caption{Data-Driven Reactivity Predictions. (A) An advanced model has been engineered for predicting pKa values, leveraging the ibond database which encompasses data from 39 distinct solvents (Adapted from ref. \cite{70}. Copyright©2020, John Wiley and Sons). The machine-learning-driven model employs Structural and Physical Organic Parameter-Based Descriptors (SPOC) to capture the molecular electronic and structural features. Training using either a neural network or the XGBoost algorithm has resulted in its remarkable predictive prowess, evidenced by a minimum mean absolute error (MAE) of 0.87 pKa units. (B) A broad-scale predictive model, developed using machine learning techniques, has been introduced (Adapted from ref. \cite{72}. Copyright©2023, John Wiley and Sons). This innovative molecular representation, designated as rSPOC, amalgamates structural, physicochemical, and solvent-related elements. The model's dataset incorporates 1115 nucleophiles, 285 electrophiles, and 22 solvents, thereby establishing it as the most expansive resource for reactivity prediction currently in existence.}
		\label{Figure 9}
	\end{figure}

	The innovative advancements in the fields of artificial intelligence and machine learning are revolutionizing how scientists analyze and comprehend immense volumes of data. These technologies merge rapid processing with unparalleled precision. Crucially, in the sphere of machine learning, the presence of comprehensive, high-caliber data is indispensable for the development of exceptional models. A monumental endeavor in this sector is the ibond database. This initiative is led by the esteemed researchers J. Cheng, S. Luo, and J. Yang from Tsinghua University, along with X. Li from Nankai University. This remarkable database aggregates activity metrics of various molecules, with a particular emphasis on parameters such as internet bond-energy database (pKa and BDE). In the year 2020, distinguished scholars J. Yang, L. Zhang, and S. Luo introduced a predictive model for pKa values, leveraging the comprehensive data from the ibond database. This data spans 39 distinctive solvents \cite{70}, as illustrated in Figure \ref{Figure 9}A. The model utilizes Structural and Physical-Organic parameter-based descriptors (SPOC), which effectively capture the electronic and structural characteristics of molecules. When this model was trained using state-of-the-art methods like neural networks or the XGBoost algorithm, it delivered exceptional performance, reflected in a Mean Absolute Error (MAE) of a mere 0.87 pKa units. Chemical reactivity, especially within polar organic reactions, is principally determined by the attributes of nucleophilicity and electrophilicity. The groundbreaking quantitative scale introduced by H. Mayr and colleagues has been instrumental in elucidating the patterns of chemical reactivity \cite{71}. By 2023, this research experienced further advancements. Notably, S. Luo and L. Zhang unveiled an enhanced predictive machine learning model \cite{72}, as detailed in Figure \ref{Figure 9}B. Termed as rSPOC, this refined molecular representation amalgamates structural, physicochemical, and solvent features. With an impressive database comprising 1,115 nucleophiles, 285 electrophiles, and 22 solvents, it is unparalleled in its capacity to predict reactivity. Impressively, when the rSPOC model was subjected to training with the Extra Trees algorithm, it exhibited remarkable accuracy in forecasting Mayr's N and E parameters, achieving $R^2$ values of 0.92 and 0.93 respectively, and maintaining a consistent MAE of 1.45 for both. This model's prowess was further evidenced when it efficiently predicted the nucleophilicity of NAD(P)H and various enamines, highlighting its potential in predicting the reactivity of previously unknown molecules. To augment its accessibility and utility, an online predictive platform was inaugurated at http://isyn.luoszgroup.com/, grounded on this pioneering model. This platform offers users the capability to predict pKa values across a wide range of X-H acidities in popularly employed solvents.
	
\begin{figure}[!t]
		\centering
		\includegraphics[width=140mm, keepaspectratio]{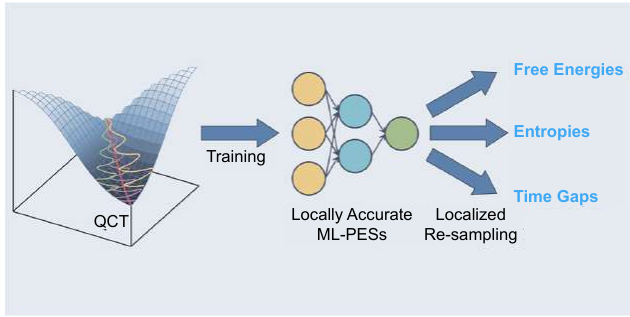}
		\caption{AI in Molecular Design. A technique known as Quasiclassical Trajectory aids in creating precise machine learning models for tracking energy changes during reactions. This was shown in two sample reactions: one involving a complex molecule's part moving and the other involving the union of two smaller molecules. These machine learning models reliably show both constant and changing aspects of reactions, including energy and entropy shifts over time and addressing time variations (Adapted from ref. \cite{91}. Copyright©2022, American Physical Society).}
		\label{Figure 11}
	\end{figure}

	The emergence of an interdisciplinary approach encourages collaboration between different disciplines. Specialists in organic chemistry, artificial intelligence and computer science, among others, are working together to create ingenious solutions to complex problems, enriching research methods and broadening the field of organic chemistry. In the complex arena of synthetic chemistry, predicting enantioselectivity in asymmetric catalysis has been a stubborn puzzle, compounded by the intricate relationship between structure and enantioselectivity \cite{83}. The challenge becomes daunting without a comprehensive understanding of synthetic space, which can discourage the exploration and discovery of asymmetric reactions \cite{84,85,86}. In 2023, X. Hong and L. Ackermann made a significant breakthrough by inventively establishing a data-driven approach that directly addresses this complex problem \cite{87}. Their tactic centred on the use of machine learning, which they combined with a deep understanding of the transition state. This combination allowed them to accurately predict the enantioselectivity of asymmetric palladium electrocatalytic C-H activation, an impressive achievement in itself. This complex process involved transforming transition state knowledge into a format that could be understood and applied by machine learning models. This modification enabled the models to perform a comprehensive evaluation of a wide range of potential scenarios, revealing unexpected effects of alkenes on enantioselectivity and providing critical insights into the dependence of the rate-determining step on olefin reactivity. A major advance in computational chemistry has been the increasing use of direct trajectory calculations, which are now an integral part of modern research \cite{88}. However, its use in the investigation of reaction mechanisms is often limited by the enormous computational cost of ab initio trajectory calculations. Against this backdrop, machine learning-based potential energy surfaces (ML-PESs) emerge as a compelling solution, offering a viable strategy to bypass these immense computational costs while maintaining the necessary precision \cite{89,90}. However, building a robust ML-PES comes with its own set of hurdles, as the training set for the potential energy surface must sufficiently cover an all-encompassing configuration space. In a landmark breakthrough in 2023, X. Hong and T. Hou charted a pioneering path \cite{91}, as shown in Figure \ref{Figure 11}. They proposed an innovative approach involving the use of quasiclassical trajectory (QCT) calculations when the required properties can be determined by localised sampling of the configuration space. This strategy facilitates the construction of locally accurate ML-PESs. Their pioneering method has been effectively demonstrated in two model reactions - the methyl migration of the i-pentane cation and the dimerisation of cyclopentadiene. Impressively, these locally precise ML-PESs showed resilience and precision in reproducing the static and dynamic features of the reactions, accurately capturing time-sensitive changes in free energy and entropy, and accounting for temporal gaps. This achievement represents a major leap forward in the advancing world of computational chemistry, and highlights the power of combining traditional computational methods with the dynamic capabilities of machine learning.
	
	\section{Conclusion}
	
	\subsection{\textit{Summary}}
		In summary, this review elucidated the crucial transformation in the field of chemistry instigated by the proliferation of automation technologies. This transition is pivotal in redirecting the focus of chemists from manual and labor-intensive tasks to intellectual inquiries as information analysts. This not only enables a more profound understanding of scientific principles but also counteracts the monotony associated with laboratory work. However, to harness the full potential of this information-centric approach, there is a compelling need to develop AI algorithms that are both predictive and capable of explainable. This necessity arises from the escalating demand for transparency in AI decision-making processes, a demand rooted in the end users' aspiration to comprehend the logic behind AI-generated outcomes instead of merely accepting them at face value. Therefore, addressing this requirement is of paramount importance for the continued development and application of AI technologies in chemistry and beyond.
	
	\subsection{\textit{Outlook}}
	
		\noindent New research paradigms are having a profound impact on organic chemistry. They are accelerating scientific discovery, reshaping interdisciplinary collaboration, provoking ethical considerations, and redefining the future role of humans in research.
\begin{itemize}
\item [(1)] 
When discussing the relationship between AI and chemistry, it's essential to consider the specific intricacies of different branches of chemistry. Researchers have made significant progress in the application of AI for chemical analysis and prediction models, as well as the development of automated chemical experimentation platforms. Through years of dedicated independent research, these achievements are now at a stage where they can be integrated to promote further advancements in the field of AI for Chemistry. The academic community is currently on the cusp of an explosion of research in this area. In the future, it is likely that the knowledge embedding and knowledge discovery techniques in scientific machine learning \cite{94,95} will be leveraged to connect current prediction models and automated experimentation platforms and generate self-evolving AI chemistry research assistant, which is illustrated in Figure \ref{Figure 12}. Here, we will focus on organic chemistry and its interplay with AI-driven advancements. Organic chemistry, which explores the structure, properties, and reactions of organic compounds and materials, is notably intricate. Its vast realm of reactions and mechanisms makes it both an exciting and challenging field. Organic molecules, with their diverse functional groups and isomers, present a rich playground for AI models. In the backdrop of AI-chemical research synergy, organic compounds' structural complexity necessitates sophisticated molecular descriptors. Graph convolutional networks (GCNs) are pivotal for processing molecular structures \cite{96}. For organic chemistry, GCNs can be particularly beneficial in recognizing structural motifs and functional groups. With the appropriate descriptors, AI can predict reactivity patterns or even the stereochemical outcomes of reactions—a historically challenging task given the 3D nature of molecules \cite{97}. Combinatorial chemistry, when applied to organic synthesis, leads to the creation of vast molecular libraries. These libraries can serve as treasure troves for drug discovery or materials science. However, the challenge here isn't just about synthesizing molecules but also about understanding their potential applications. Organic molecules often display biological activity due to their interactions with biomolecules, and predicting these interactions requires understanding the subtle interplays of molecular shapes, electronics, and more. The idea of knowledge discovery in scientific machine learning offers unique prospects for organic chemistry \cite{98,99}. Reaction mechanisms, which are central to organic chemistry, often involve intricate pathways with intermediates, transition states, and competing reactions. Traditional methods of determining these mechanisms involve kinetic studies, isotope labeling, and more. However, if symbolic mathematics, combined with AI, can help elucidate these pathways, it would revolutionize our understanding and teaching of organic reactions. Knowledge embedding, as mentioned, becomes even more pertinent from an organic chemist's perspective. Organic chemistry has a wealth of heuristic rules, from Markovnikov's rule for electrophilic addition to Baldwin's rules for ring closures. If AI models can be instilled with this embedded knowledge, their predictions wouldn't just be data-driven but would also align with chemists' intuitive understanding, offering a deeper and more insightful perspective.
			\begin{figure}[!t]
			\centering
			\includegraphics[width=140mm, keepaspectratio]{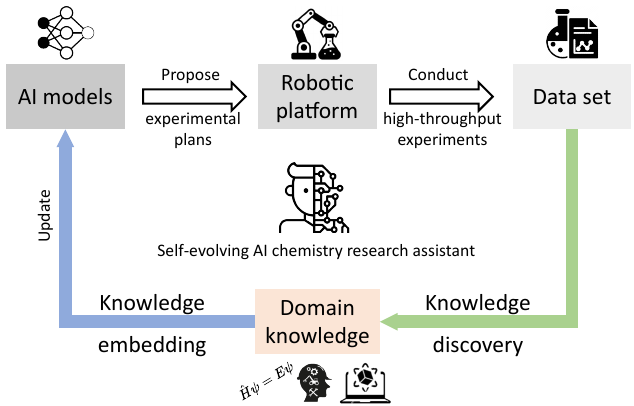}
			\caption{Pipeline of generating self-evolving AI chemistry research assistant.}
			\label{Figure 12}
			\end{figure}
\item [(2)] 
The world of artificial intelligence (AI) has seen a considerable shift in the past few years, with general-purpose models like the Generative Pre-trained Transformer (GPT) gaining immense popularity. This evolution is not isolated to linguistics; it has spilled over to other specialized domains, most notably chemistry. And as we dive deeper into this confluence of chemistry and AI, it's worth noting the strides we've made, especially in terms of chemical big models and language models. ChEMBL, for instance, stands out as a beacon in this realm. As a manually curated database of bioactive molecules, its value is magnified when combined with transformer models. By training these models on the ChEMBL dataset, researchers have been able to predict a plethora of molecular properties and activities, bridging the gap between AI's potential and practical chemistry \cite{79}. Further, the integration of the SMILES notation - Simplified Molecular Input Line Entry System - demonstrates another innovative step. With SMILES, chemical structures get a textual representation, allowing for easier integration into language models. This integration has enabled these models to predict, generate, and optimize molecular structures, a feat previously deemed far-fetched. Then there's ChemBERTa, reminiscent of the BERT model for natural languages but tailor-made for the world of chemistry. As a transformer-based model, ChemBERTa taps into vast amounts of chemical data, and upon fine-tuning, serves purposes ranging from property prediction to the complex realm of retrosynthesis \cite{80}. Speaking of retrosynthesis, the RETRO model stands out in this sphere. Trained on vast databases of chemical reactions, RETRO predicts potential retrosynthetic pathways, offering insights into potential precursors and reactions for target molecules \cite{81}. Beyond tools like RETRO, even large language models such as BERT and GPT have been applied in inverse synthesis, where, given a product, the model suggests potential synthetic routes and reactants. However, it's not merely the applications of neural networks in AI chemistry that are evolving — it's the very essence of their nature. The chronology of AI saw a humble beginning with simple neural network architectures, but today, we witness behemoths like GPT-4, armed with billions of parameters \cite{82}. This scaling up in AI chemistry brings about an interesting conundrum, one that juxtaposes model complexity with interpretability. There's an undeniable allure to the capabilities of larger models. Yet, in a domain like chemistry, where the stakes are high and the margin for error is minuscule, the need for interpretability is undeniable. Reflecting upon AI's role in chemistry, it becomes clear that interpretability, though significant, isn't the lone beacon guiding this ship. The overarching vision for AI in chemistry is vast and multi-dimensional: Prediction is at the forefront, with sophisticated models promising unparalleled accuracy in forecasting chemical properties and reactions. Exploration is another frontier, with AI providing a lens to navigate the intricate maze of chemical compounds and reactions, some of which might elude human intuition. Optimization presents the promise of revolutionizing chemical processes by pushing the boundaries to achieve greener and more efficient synthesis methodologies. Ultimately, while the initial attraction towards AI chemistry might have been its promise of interpretability, the reality is far more nuanced. The intersection of AI and chemistry is a dynamic space, ever-evolving and constantly being reshaped by the challenges and needs intrinsic to the chemical domain. As we continue to traverse this journey, it's evident that AI's contribution to chemistry is bound to be as transformative as it is multifaceted.
\item [(3)] 
The emergence of new paradigms in artificial intelligence and automation has also raised significant ethical concerns. These technologies are permeating various research fields, stirring debates about data privacy and the transparency of AI algorithms. In particular, the question arises: how can we leverage vast amounts of data without compromising privacy? One potential solution is Federated Learning. Federated Learning is an approach that allows for model training across multiple decentralized devices or servers holding local data samples, without data exchange \cite{92,93}. This not only preserves data privacy but also facilitates the collaborative nature of AI without centralizing sensitive information. Within the context of organic chemistry, this means that institutions or labs can contribute to the development and training of shared models without ever exposing individual data points, ensuring that sensitive or proprietary chemical data remains confidential. Furthermore, transparency in AI algorithms can be enhanced by incorporating explainable AI (XAI) techniques, which aim to make the inner workings of models more interpretable for human researchers. As the discipline of organic chemistry progresses, integrating such techniques can offer insights into why a particular AI-driven prediction or recommendation was made, bridging the gap between human intuition and machine learning. As the landscape of organic chemistry transforms under AI's influence, the above considerations, combined with federated learning, will play a pivotal role. While AI performs routine tasks, allowing human researchers to focus on conceptual and creative endeavors, it's essential to harness AI tools responsibly and ethically. The union of humans and AI in research should be symbiotic, with each enhancing the other's capabilities. For now, as we stand on the brink of a new epoch in organic chemistry, scientists need not fear AI replacing them. Instead, the emphasis should be on adapting and mastering AI tools, ensuring that human unique skills continue to complement AI, making the collaboration more powerful than either could achieve alone.
			
\end{itemize}
		\subsection{\textit{Recommendations}}
		\noindent The rise of interdisciplinary research is revolutionizing scientific inquiry. By fusing organic chemistry, computer science, and artificial intelligence, new systems now predict reaction outcomes and optimize synthetic routes. This melding has significantly broadened the scope of organic chemistry. Within the evolving panorama of synthetic platforms, microfluidics and flow chemistry are particularly striking. They offer meticulous control over reaction conditions, enhance reproducibility, and efficiently scale reactions. The advent of robot-assisted synthesis, as demonstrated by systems like ChemSpeed and innovations from pharmaceutical companies, utilizes robots to perform repetitive tasks, ensuring consistent conditions, reducing human errors, and amplifying throughput \cite{78}. Concurrently, 3D printing in organic synthesis serves as an innovative tool for creating custom reactionware, thereby enhancing experimental design versatility. Another significant advancement is in situ monitoring techniques. Devices like ReactIR and NMR, which offer real-time reaction monitoring, have become crucial in understanding reaction kinetics, identifying intermediates, and determining reaction endpoints. With the scientific community gravitating towards sustainability, platforms emphasizing renewable resources and green reaction conditions have risen to prominence. For researchers venturing into this field, the ability is multifaceted, as follows:
		
		• Keeping abreast of recent literature, conferences, and workshops on automation and AI in chemistry is vital.
		
		• Collaborating with professionals like computer scientists, engineers, and data analysts can offer fresh insights.
		
		• A basic understanding of programming, particularly Python due to its cheminformatics libraries, is indispensable.
		
		• Starting with small-scale automation projects can offer familiarity with the domain before scaling up.
		
		• Regardless of the project size, safety is paramount, especially considering the unpredictable challenges posed by automation.
		
		When exploring machine learning predictive models for organic synthesis, a solid foundation is crucial. While these models are powerful, they require precise data that encompasses physicochemical properties, spectral details, and known synthetic pathways. Leveraging existing data sources, such as literature, patents, and academic texts, can prevent redundant experimental efforts and optimize resources. It is vital to integrate diverse datasets; correlating spectral data with reaction outcomes can shed light on mechanisms or reveal unforeseen by-products. Although large datasets are tempting, the focus should be on data quality; subpar data can lead to inaccurate predictions, highlighting the need for careful data curation and validation. A fascinating aspect of machine learning is transfer learning, wherein models trained on existing datasets can be refined with specific experimental data, optimizing the learning path and enhancing model robustness. Furthermore, scholars with a deep understanding of two or more fields often find themselves more inspired and creative, showcasing the unique allure and excitement of interdisciplinary science. For those eager to build an automation system in the laboratory, proficiency in electrical knowledge, simple mechanical structure design, CAD software, robot programming and control technology, programmable logic controllers (PLC), and communication technology is essential. Additionally, a strong hands-on ability is crucial. Although professional companies can be engaged if the budget permits, the cost-effectiveness and subsequent maintenance and transformation pose challenges. Given the limited resources of scientific research groups, deploying artificial intelligence and automation systems requires substantial resources. A practical solution is to collaborate with the industry on scientific issues of mutual interest. The industry may have a wealth of data and application scenarios in specific fields, while the academic community possesses the intellectual resources to identify and address scientific problems. Such collaboration can promote progress in the field and facilitate the sharing of scientific research results.



\end{document}